\documentclass[lettersize,journal]{IEEEtran}
\usepackage{amsmath,amsfonts}
\usepackage{array}
\usepackage[caption=false,font=normalsize,labelfont=sf,textfont=sf]{subfig}
\usepackage{textcomp}
\usepackage{stfloats}
\usepackage{url}
\usepackage{verbatim}
\usepackage{graphicx}
\usepackage{cite}
\usepackage{pifont}
\usepackage{graphicx}
\usepackage{hyperref}
\usepackage{svg}      
\usepackage{subcaption}
\usepackage{subfloat}
\usepackage{booktabs}
\usepackage{xcolor}
\usepackage{multirow}
\usepackage{rotating} 
\usepackage{booktabs} 
\usepackage{amssymb}

\usepackage{amsmath,graphicx}
\usepackage{algorithm}
\usepackage{algpseudocode}
\usepackage{multirow}

\usepackage{amsmath}

\begin{document}

\setlength{\abovedisplayskip}{1pt}
\setlength{\belowdisplayskip}{1pt}
\setlength{\floatsep}{3pt plus 1.0pt minus 1.0pt}
\setlength{\intextsep}{3pt plus 1.0pt minus 1.0pt}
\setlength{\textfloatsep}{3pt plus 1.0pt minus 1.0pt}
\setlength{\parskip}{1pt}
%
\title{SignalLLM: A General-Purpose LLM Agent Framework for Automated Signal Processing}
%

%
%

\author{Junlong~Ke\textsuperscript{*},
        Qiying~Hu\textsuperscript{*},
        Shenghai~Yuan,
        Yuecong~Xu,
        and~Jianfei~Yang,~\IEEEmembership{Senior Member,~IEEE}

\thanks{*~Junlong~Ke and Qiying~Hu contributed equally to this work (kjl25@mails.tsinghua.edu.cn, huqy24@mails.tsinghua.edu.cn). The work was carried out during their undergraduate research internship at the MARS Lab, Nanyang Technological University, Singapore.}
\thanks{Junlong~Ke and Qiying~Hu are with the Department of Electronic Engineering, Tsinghua University.}
\thanks{Shenghai~Yuan and Yuecong~Xu are with the School of Electrical and Electronic Engineering, Nanyang Technological University, Singapore.}
\thanks{Jianfei~Yang is with the MARS Lab in the School of Mechanical and Aerospace Engineering and the School of Electrical and Electronic Engineering, Nanyang Technological University, Singapore.}
\thanks{Jianfei~Yang is the corresponding author (jianfei.yang@ntu.edu.sg).}
}

%
%

\maketitle

\begin{abstract}

Modern signal processing (SP) pipelines, whether model-based or data-driven, often constrained by complex and fragmented workflow, rely heavily on expert knowledge and manual engineering, and struggle with adaptability and generalization under limited data. In contrast, Large Language Models (LLMs) offer strong reasoning capabilities, broad general-purpose knowledge, in-context learning, and cross-modal transfer abilities, positioning them as powerful tools for automating and generalizing SP workflows. Motivated by these potentials, we introduce SignalLLM, the first general-purpose LLM-based agent framework for general SP tasks. Unlike prior LLM-based SP approaches that are limited to narrow applications or tricky prompting, SignalLLM introduces a principled, modular architecture. It decomposes high-level SP goals into structured subtasks via in-context learning and domain-specific retrieval, followed by hierarchical planning through adaptive retrieval-augmented generation (RAG) and refinement; these subtasks are then executed through prompt-based reasoning, cross-modal reasoning, code synthesis, model invocation, or data-driven LLM-assisted modeling. Its generalizable design enables the flexible selection of problem solving strategies across different signal modalities, task types, and data conditions. We demonstrate the versatility and effectiveness of SignalLLM through five representative tasks in communication and sensing, such as radar target detection, human activity recognition, and text compression. Experimental results show superior performance over traditional and existing LLM-based methods, particularly in few-shot and zero-shot settings. Our codes are available at https://github.com/uestc1209radar/SignalLLM.

\end{abstract}

\begin{IEEEkeywords}
Signal processing, large language models, agentic AI.
\end{IEEEkeywords}

\IEEEpeerreviewmaketitle

\section{Introduction}\label{intro}

\IEEEPARstart{S}{ignal} processing (SP) underpins wireless communications, image analysis, radar and sonar systems, and the Internet of Things (IoT)~\cite{lu2023edge}. Traditional approaches are dominated by model-based and data-driven methods, both facing certain limitations~\cite{mansouri2020data}. Model-based pipelines require extensive expertise, complex workflows, and labor-intensive development, while data-driven approaches often overfit and generalize poorly, despite large annotated datasets and computational resources~\cite{salau2022recent}. Recently, Large Language Models (LLMs), such as ChatGPT, have shown remarkable abilities in natural language understanding, image editing, summarization, and generation~\cite{chang2024survey, wen2025stop, wen2025token}. With large-scale pretraining and massive parameterization, LLMs capture broad knowledge and perform complex multimodal reasoning. Motivated by their generalizability, emerging research explores their potential for SP: Shen \textit{et al.}\cite{shen2025gpiot} applied LLMs to automate code generation, An \textit{et al.}\cite{an2024iot} designed prompts for zero- and few-shot IoT reasoning, and Wu \textit{et al.}~\cite{wu2024netllm} adapted pre-trained LLMs for various networking tasks, considerably surpassing conventional neural networks.

Although powerful and promising, current LLM-based methods for signal processing (SP) remain inadequate in supporting practical applications: (1) most existing approaches are tailored to specific steps of individual SP tasks and are often developed and evaluated on narrowly scoped datasets, lacking a framework capable of addressing multiple, complex SP tasks; (2) when planning for specific SP tasks, LLMs tend to offer narrow, suboptimal solutions due to the limited presence of SP-related knowledge and datasets in the pretraining corpora, which leads LLMs to generate generic outputs rather than domain-specific responses; (3) current LLM-based approaches fail to accommodate the diversity of solution paradigms required under different task constraints. In practice, various SP tasks require different solution strategies: code generation with external compilers~\cite{shen2025gpiot} is suitable for computationally intensive problems, LLM/MLLM-based SP task reasoning~\cite{an2024iot,ji2024hargpt} excels in zero- or few-shot scenarios, and LLM-assisted SP task modeling~\cite{wu2024netllm} can facilitate parameter transfer to improve generalization and robustness. However, existing methods typically adopt a single, fixed solution type without adapting to task-specific constraints, resulting in suboptimal performance.

To address the issues outlined above, we conduct three focused analyses. For the first limitation, advances in agentic LLM systems for tasks such as mathematical problem solving~\cite{romera2024mathematical}, web navigation~\cite{nakano2021webgpt}, and medical decision-making~\cite{li2024agent} highlight a promising paradigm. Inspired by these developments, we argue that signal processing similarly requires an agentic framework that integrates structured planning with adaptive execution, allowing LLMs to solve complex tasks through a coordinated two-stage process. Second, current LLMs struggle with specialized SP tasks due to limited reasoning ability and the lack of domain-specific training. This motivates a planning stage that incorporates task decomposition, targeted dataset construction, and retrieval-augmented reasoning to generate context-aware solutions. Third, existing approaches often rely on a single, fixed solution path~\cite{shen2024hugginggpt,an2024iot,wu2024netllm} and fail to account for variations in task complexity and resource constraints. We aim to comprehensively characterize the landscape of LLM-based approaches for signal processing by constructing a taxonomy that encompasses code generation, LLM and MLLM reasoning, LLM-assisted modeling, and optimization. For each method, we analyze its strengths, limitations, and applicability to guide adaptive strategy selection. To enable flexible execution, we further introduce a refinement module that allows memory-guided agents to dynamically select among diverse tools, models, and solvers based on task-specific requirements.

Leveraging the above insights, we propose SignalLLM, a general-purpose agentic framework designed to address a wide range of complex SP tasks. It integrates high-level planning with adaptive execution through five tightly coupled components. First, a task decomposition module employs domain-specific retrieval and in-context learning to translate high-level SP requirements into coherent subtask objectives. Second, a planning module with adaptive RAG generates subtask-level solutions. which are further improved by a dedicated refinement component to support diverse solution paradigms. To overcome the rigidity of single-paradigm execution, SignalLLM incorporates two complementary modules: (1) an LLM-assisted reasoning module enabling prompt engineering, code generation with external compilers, and cross-modal understanding, and (2) an LLM-assisted modeling module supporting optimization and parameter transfer. Together, these components form a unified pipeline that addresses core limitations of existing LLM-for-SP methods. We validate SignalLLM on diverse SP tasks spanning transmission, recognition, and perception, under challenging constraints such as few-shot radar target detection and zero-shot human activity recognition, highlighting its adaptability and generalization across modalities and domains.

The contributions of this paper are summarized as follows:

\begin{itemize}
\item In light of persistent challenges such as heavy dependence on expert knowledge, limited adaptability to diverse data conditions, and the absence of a general-purpose and flexible LLM-based SP solution framework, we propose SignalLLM, the first agentic LLM framework specifically designed to address complex SP tasks with structured planning and hybrid execution strategies.

\item SignalLLM stands out for its automated and adaptable workflow, which intelligently selects the most suitable solution from a range of paradigms. To the best of our knowledge, this is the first framework that both (i) systematically summarizes the LLM-for-SP landscape through a functional taxonomy and (ii) leverages agentic workflows to dynamically compose solutions from a diverse SP functional taxonomy, often discovering strategies that surpass human-designed heuristics.

\item We perform extensive evaluations of SignalLLM across five representative SP tasks, including text data transmission, few-shot radar target detection, and zero-shot human activity recognition. The results demonstrate that SignalLLM consistently exceeds both traditional SP methods and previous agent-based approaches, particularly in scenarios characterized by limited data availability. 
\end{itemize}
\section{Related Work} 

\subsection{Large Language Model Powered SP Tasks}
Current research on integrating Large Language Models (LLMs) into signal processing (SP) tasks generally falls into two overarching paradigms. 

The first paradigm leverages LLMs as intelligent interfaces or coordinators within SP systems, especially in the context of the Internet of Things (IoT) and broader wireless network environments. In this process, the input and output of LLMs are not necessarily redefined; rather, they continue to function as agents employing natural language. In such settings, LLMs are employed to interpret sensor data, generate code for SP tasks, and facilitate natural human-machine interactions. Recent developments, such as Penetrative AI~\cite{xu-etal-2024-penetrative} and HarGPT~\cite{ji2024hargpt}, push this paradigm further by directly ingesting raw signal data and utilizing chain-of-thought prompting to enable complex reasoning over physical-world IoT tasks. Building upon this, IoT-LLM~\cite{an2024iot} enhances the reasoning capabilities of LLMs in physical environments by integrating structured sensor data, commonsense prompting, and retrieval-augmented knowledge. GPIOT~\cite{shen2025gpiot} demonstrates how LLMs, equipped with code interpretation capabilities, can be used to autonomously generate code for performing IoT tasks in distributed system.

The second paradigm involves directly fine-tuning LLMs on SP tasks, capitalizing on their potential for general pattern recognition to replace traditional data-driven methods. In this process, the modeling capabilities of LLMs are utilized, and their inherent parametric pattern recognition abilities are either leveraged or directly incorporated as part of the modeling framework. For instance, NetLLM~\cite{wu2024netllm} adapts pre-trained LLMs to networking tasks by introducing multimodal encoders and task-specific output heads for processing sequential, tabular, and graph-structured data. 

Despite recent progress, as noted in Section~\ref{intro}, a dedicated framework that enables LLMs to deliver tailored and adaptable solutions across diverse signal processing (SP) tasks is still lacking. Existing approaches often focus on isolated subtasks, rely on fixed solution paths, or lack the capacity to dynamically adapt across modalities, constraints, and objectives. Addressing this shortcoming is essential to unlock the full potential of LLMs in SP applications.

\subsection{Agentic AI}
The emergence of Agentic AI~\cite{sapkota2025ai} marks a shift from single-agent systems toward collaborative intelligence, where multiple specialized agents coordinate to solve complex tasks through goal decomposition and dynamic interaction~\cite{yao2023react}. Unlike conventional AI agents that operate in isolation with fixed tool-based routines, Agentic AI systems incorporate foundation models (e.g., LLMs, LIMs), persistent memory, and decentralized planning mechanisms. This architecture enables adaptive, context-aware behavior beyond the scope of traditional automation. In signal processing (SP), such agentic frameworks offer transformative capabilities for managing complex, multi-stage pipelines, enabling not only structured, step-by-step task planning, but also the dynamic selection and composition of optimal subtask-level solutions tailored to specific objectives, data modalities, and system constraints.

\section{Methodology}

\begin{figure*}[t]
\centering
    \includegraphics[width=1\textwidth]{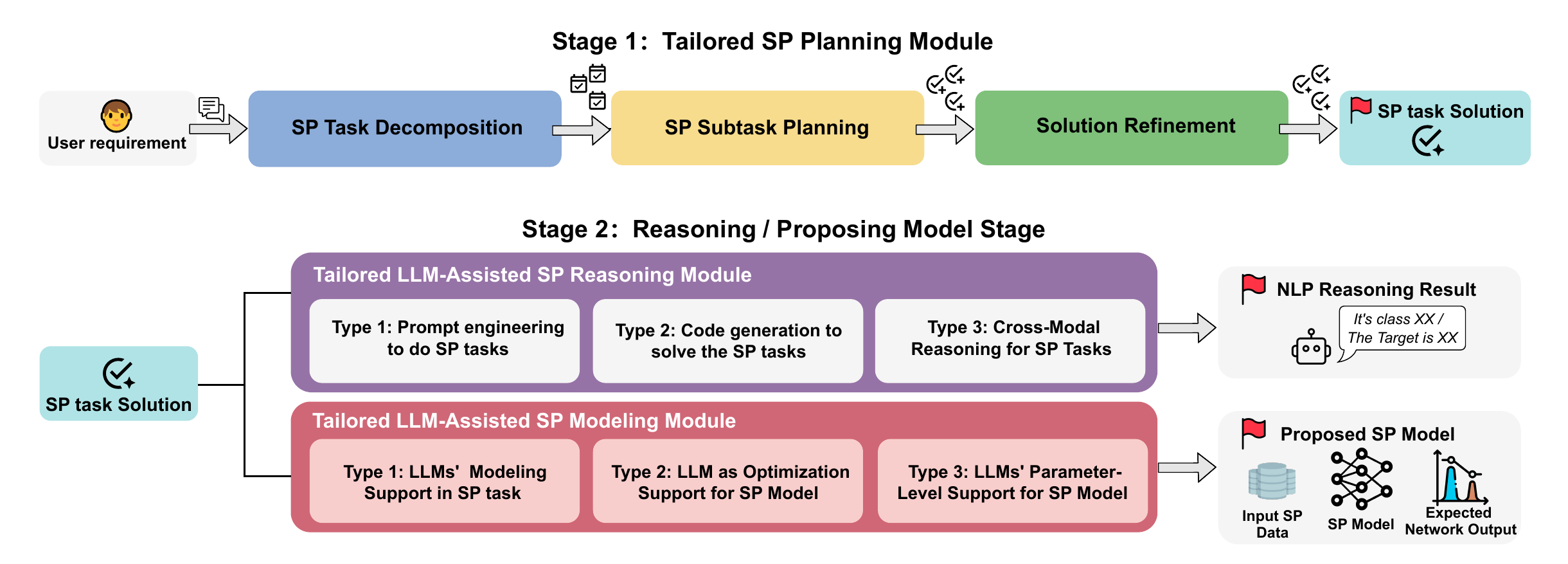}

\caption{Overview of the SignalLLM framework. SignalLLM operates in two main stages. Stage 1 focuses on tailored planning, where user requests are processed through task decomposition, subtask planning, and solution refinement. Stage 2 is dedicated to execution, carrying out the plan using either the LLM-Assisted SP Reasoning Module for reasoning-based tasks or the LLM-Assisted SP Modeling Module for tasks requiring LLM-assisted model creation.} 
    \label{SignalLLM}
\end{figure*}

We propose SignalLLM, a general-purpose framework designed to tackle complex SP tasks with targeted solutions. SignalLLM comprises five core components, as illustrated in Figure~\ref{SignalLLM}.
Initially, SignalLLM plans and deconstructs a task using its \textit{SP Task Decomposition}, \textit{SP Subtask Planning}, and \textit{Solution Refinement} modules. The resulting subtasks are then classified and routed to one of two specialized pathways. Reasoning-based subtasks are handled by a \textit{LLM-Assisted SP Reasoning Module}, while those requiring LLM-assisted model creation are processed by a  \textit{LLM-Assisted SP Modeling Module}. Finally, the framework produces the appropriate output, such as a specialized SP model or a natural language response, tailored to the initial requirements.
\subsection{Tailored SP Planning Module} 

\subsubsection{SP Task Decomposition}
\begin{figure}[t]
\centering
\includegraphics[width=3.5in]{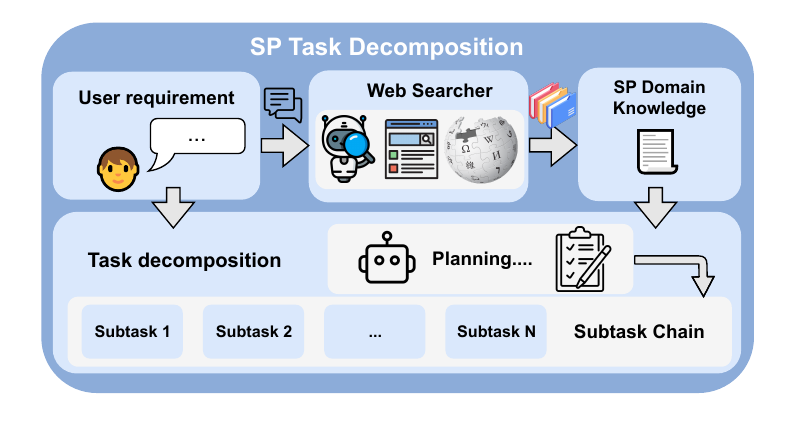}

  \caption{The illustration of {SP Task Decomposition} Module. A user's natural language request is processed by a Web Searcher to acquire SP Domain Knowledge. which is then used to decompose the high-level goal into a structured subtask chain.} 
\label{Decomposition}
\end{figure}

As shown in Figure~\ref{Decomposition}, for the initial SP task requirements expressed in natural language, we adopt the Toolformer framework~\cite{schick2023toolformer} to construct a Web Searcher that leverages web search APIs to build task-specific SP Domain Knowledge. Subsequently, SignalLLM utilizes in-context learning to decompose these requirements into a series of subtask plans. Through the task decomposition module, a complex signal processing (SP) task can be broken down into simpler sub-steps and progressively completed.

\subsubsection{SP Subtask Planning}
\begin{figure}[t]
\centering
\includegraphics[width=3in, trim={0.5cm 0.2cm 0.5cm 0.2cm}, clip]{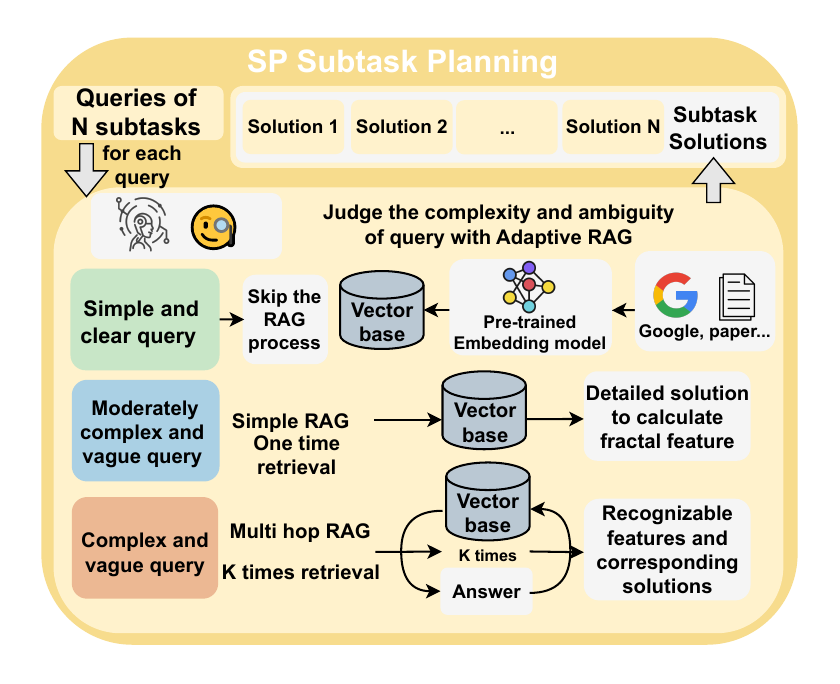}
  \caption{The illustration of SP Task Solution Planning Module. This module integrates complexity-aware RAG mechanisms with agent-based planning to dynamically support LLMs in solving SP subtasks.} 
\label{planning}
\end{figure}

Although the SP Task Decomposition module provides a preliminary decomposed plan, the steps are often abstract rather than concrete, and some step sequences may be illogical, leading to instability in subsequent agent operations. As shown in Figure~\ref{planning}, to address this, we have designed several modules to further construct specific, agent-friendly subtask solutions.

In the SP Subtask Chain Construction module, subtasks areorganized sequentially, structured according to their logical order and dependencies. This preliminary structuring ensures efficient execution and minimizes redundancy.

LLMs exhibit varying levels of competence when dealing with different types of SP subtasks. For straightforward subtasks, such as basic transformations or filtering, LLMs can often generate accurate solutions directly. However, for more complex or domain-specific tasks (e.g., interpreting ambiguous signal features or designing wireless sensing schemes), LLMs often lack sufficient domain knowledge and struggle to reason over underspecified problem descriptions. In such cases, retrieval-augmented generation (RAG) is essential to enhance the LLM's ability to understand and plan for unfamiliar subtasks. To address this issue, we introduce the SP Task Solution Planning Module, which integrates complexity-aware RAG mechanisms with agent-based planning to dynamically support LLMs in solving SP subtasks. The module follows a hierarchical structure based on subtask complexity and ambiguity, as assessed using the method from~\cite{mallen2022not}, and operates as follows:
\begin{itemize}
    \item For simple and well-defined subtasks, the LLM independently generates solutions without external assistance. This is applicable when the subtask falls within the LLM's parametric knowledge.
    \item When the subtask exhibits moderate complexity or partial ambiguity, a single-round retrieval mechanism is employed to supplement the LLM with domain-specific background information. Formally, given a subtask plan $\boldsymbol{q}$, relevant contextual support $\boldsymbol{s}$ is retrieved as $\boldsymbol{s}=\operatorname{Retriever}(\boldsymbol{q}, \boldsymbol{v})$, where $\boldsymbol{v}$ denotes the external vector index. The retrieved content is concatenated with the plan and processed jointly by the LLM to yield an informed response.
    \item For subtasks characterized by high complexity or compounded ambiguity, a multi-hop RAG strategy~\cite{khot2022decomposed} is employed to iteratively construct a coherent solution. At iteration $i$, contextual information $\boldsymbol{c}_{i}$ is used in conjunction with $\boldsymbol{q}$ to retrieve additional evidence $\boldsymbol{s_i}=\operatorname{Retriever}(\boldsymbol{q}, \boldsymbol{c}_{i}, \boldsymbol{v})$. The LLM then generates an intermediate output $\boldsymbol{a}_{i}=\operatorname{LLM}\left(\boldsymbol{q}, \boldsymbol{c}_{i}, \boldsymbol{s}_i\right)$, which is incorporated into the evolving subtask context:
    \begin{equation}
        \boldsymbol{c}_{i+1}=\left(\boldsymbol{d}_{1}, \ldots, \boldsymbol{d}_{i}, \boldsymbol{a}_{1}, \ldots, \boldsymbol{a}_{i}\right) .
    \end{equation}
\end{itemize}

This iterative process continues until a stable and contextually grounded solution is obtained. By selectively engaging external knowledge sources according to subtask complexity, the proposed planning module effectively enhances the interpretability and problem-solving robustness of LLMs, particularly in domains where domain-specific procedural reasoning is indispensable.

\subsubsection{Solution Refining Module}

Large language models (LLMs) are often used to generate code for SP tasks by replicating traditional model-based or data-driven solutions. However, such approaches primarily focus on algorithm reproduction through code synthesis, without fully leveraging the intrinsic problem-solving capabilities of LLMs. In practice, various SP tasks require different solution strategies: code generation of traditional SP solutions with external compilers~\cite{shen2025gpiot} is suitable for computationally intensive problems, but often fall short in flexibility, generalizability, and data scarcity scenarios. In contrast, LLM/MLLM-based SP task reasoning~\cite{an2024iot,ji2024hargpt} excels in zero- or few-shot scenarios, and LLM-assisted SP task modeling~\cite{wu2024netllm} can facilitate parameter transfer for improved generalization and performance. We further design the Solution Refining Module, establish agent memory that stores possible LLM-for-SP solutions along with their respective strengths and limitations. A refinement agent then evaluates these alternatives in comparison to the original solution, enabling a comprehensive analysis and selection of the most appropriate approach.
\subsection{Tailored LLM-Assisted SP Reasoning Module} 

\begin{figure*}[t]
\centering
\includegraphics[width=7in]{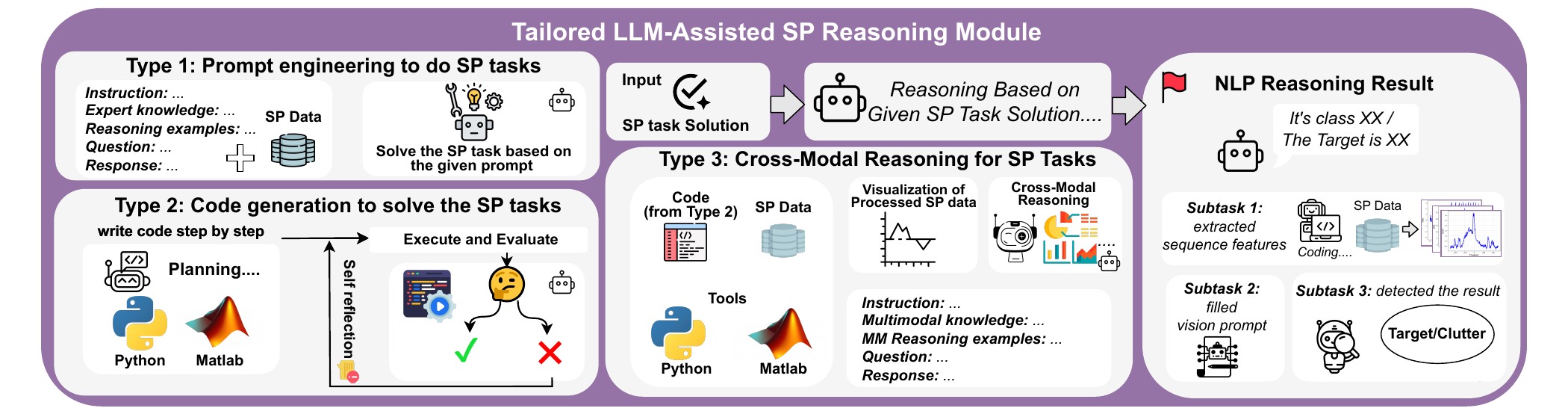}
\caption{The illustration of the proposed Tailored LLM-Assisted SP Reasoning Module and its three operational modes. (1) Prompt engineering to do SP tasks: Employs direct task execution through sophisticated prompt design. (2) Code generation to solve the SP tasks: Generates and refines Python or Matlab code via planning and self-reflection. (3) Cross-Modal Reasoning for SP Tasks: Integrates code-generated data with visual representations to solve complex multimodal tasks.}
\label{Reasoning}
\end{figure*}

As shown in Figure~\ref{Reasoning}, for SP tasks that are suitable for pure agent-based solutions, we refer to them as LLM agent-based SP tasks. To address these tasks, SignalLLM incorporates three types of reasoning modules, including Prompt Engineering for SP Tasks, Code Generation for SP Tasks and Cross-Modal Reasoning for SP Tasks. Techniques such as chain-of-thought prompting or few-shot prompting~\cite{sahoo2024systematic} are utilized to improve the accuracy and relevance of SignalLLM-generated responses.

\subsubsection{Prompt Engineering for SP Tasks}
Prompt engineering serves as a foundational reasoning mechanism in SignalLLM, enabling systematic integration of domain expertise into language model operations. 
For the simplest reasoning tasks, we directly provide the LLM agent with structured prompts comprising \textit{Instruction}, \textit{Expert Knowledge}, \textit{Reasoning Examples}, \textit{Question}, and \textit{Response Format}. This allows the LLM agent to solve the SP task based on the given prompt, leveraging its in-context learning capabilities.

\subsubsection{Code Generation for SP Tasks}
To bridge the semantic gap between natural language specifications and executable SP implementations while preserving mathematical rigor, this module leverages the code generation capabilities of LLMs. Built upon the Toolformer framework~\cite{schick2023toolformer}, SignalLLM addresses SP subtasks by generating functional code through chain-of-thought reasoning and self-reflection~\cite{wang2024executable}, guided by subtask plans and specific solution requirements. To ensure computational precision, external software tools such as Python and MATLAB are invoked as solvers to execute complex calculations and return accurate results.

\subsubsection{Cross-Modal Reasoning for SP Tasks}
Extending the code generation framework, this module establishes a multimodal reasoning pipeline that synchronizes textual instructions, mathematical formulations, and visual representations. %
The input to this module includes \textit{Instruction}, \textit{Multimodal Knowledge}, \textit{MM Reasoning Examples}, \textit{Question}, and \textit{Response Format}. By leveraging cross-modal reasoning, SignalLLM can interpret and analyze multimodal inputs, such as visual plots and textual data, to provide more comprehensive and accurate solutions. This approach is particularly effective for tasks that require the integration of multiple data modalities, as demonstrated in recent studies~\cite{yoon2024my}.  

\subsection{Tailored LLM-Assisted SP Modeling Module} 

\begin{figure*}[t]
\centering
\includegraphics[width=6in]{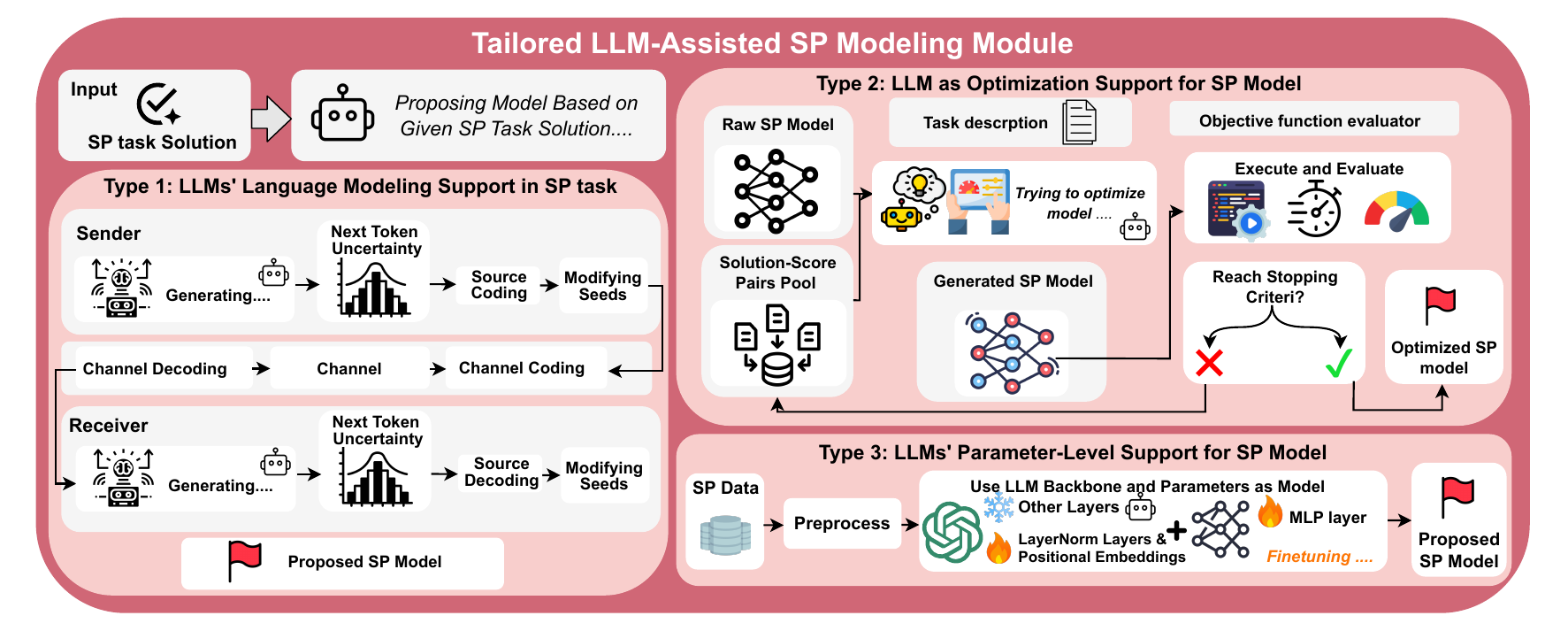}
\caption{The illustration of the proposed Tailored LLM-Assisted SP Modeling Module and its three approaches to data-driven modeling. (1) LLMs' Language Modeling Support in SP task: Directly utilizes the LLM's own language modeling capabilities for tasks such as semantic communication. (2) LLM as Optimization Support for SP Model: Employs the LLM to tune the hyperparameters of an external SP model. (3)  LLMs' Parameter-Level Support for SP Model: Leverages the LLM's pre-trained parameters to initialize and fine-tune new SP models via knowledge transfer.}

\label{LLMMG}
\end{figure*}

Traditional NLP question-answering agent approaches often fall short when applied to SP tasks that require the construction of complex input-output systems rather than simple pattern recognition. As shown in Figure~\ref{LLMMG}, to address this limitation, we propose a data-driven SP framework that operates at three distinct levels: model-level, parameter-level, and optimization-level that constructing the proposed Tailored SP LLM-Assisted Modeling Module. More precisely, SignalLLM incorporates three types of data-driven signal processing modules, including LLM as SP Task Modeling, LLM as an Optimizer and Parameter Transfer from Pre-Trained LLM. This framework leverages the statistical patterns and knowledge embedded in LLMs during their pre-training phase to optimize SP tasks. The proposed approach is particularly effective in resource-constrained scenarios where minimal retraining or fine-tuning is required, making it a versatile solution for a wide range of SP applications.

\subsubsection{LLM as SP Task Modeling (LLM-Supported Modeling)}
The rise of LLMs marks a major milestone in natural language processing, with capabilities closely aligned to human cognition. In modern communication systems, which are shifting toward semantic communication for improved spectral efficiency, LLMs remain underutilized at the system level of SP. SignalLLM, as illustrated in Figure~\ref{LLMMG}, leverages LLMs’ pretrained linguistic modeling to directly perform SP tasks without task-specific fine-tuning. By embedding LLMs into natural language coding frameworks (Algorithm~\ref{LLM4code}), SignalLLM models source coding as a probabilistic inference task, using next-token uncertainty to guide symbol distribution, which significantly streamlines encoding efficiency.
\begin{algorithm}
\label{algorithm1}
    \caption{SignalLLM assisted source coding and decoding} \label{LLM4code}
    \scriptsize
    \begin{algorithmic}[1] 
        \Require Text string $\mathcal{T}$ to be transmitted with $N$ tokens; Index position set $\mathcal{S}=\emptyset$; The prior context length $K$; Prior context matrix $P=\emptyset$;
        \Ensure The recovered text string $\mathcal{T}^{\prime}$, $\mathcal{T}^{\prime}=\emptyset$ initially.
        \State Divide $\mathcal{T}$ into tokens through LLM
        \For{$i \gets 1$ to $N$}
        \State Find the prior $K$ length context $P_i$ before transmitted token $i$, $P=[P; P_i]$
        \EndFor 
        \State Using the LLM and matrix $P$, predict the probability distribution over the entire token vocabulary for tokens in positions 1 to $N$, and rank all tokens in each position from highest to lowest probability.
        \State Retrieve the index position $S_i$ of transmitted token $i$ in all ranked tokens, $\mathcal{S}=\mathcal{S} \cup S_i$ (if the transmitted token is at the top position, its value is 0, with subsequent positions increasing sequentially)
        \State Encode the redundant set $\mathcal{S}$ using entropy coding method.
        \State Decode the bitstream to gain decoded set $\mathcal{S}^{\prime}$ using entropy decoding method, containing $N^{\prime}$ index positions
         \For{$i \gets 1$ to $N^{\prime}$}
        \State Predict probability distribution over the entire token vocabulary in the token $i$ place through LLM, rank all tokens from highest to lowest probability
        \State Obtain decoded token $T_i^{\prime}$ through the recovered index position, $\mathcal{T}^{\prime}=\mathcal{T}^{\prime} \cup T_i^{\prime}$ 
        \EndFor 
        \State \Return The recovered text string $\mathcal{T}^{\prime}$
    \end{algorithmic}
\end{algorithm}

\subsubsection{LLM as an Optimizer (LLM as Optimization Support for SP Model)}

Optimizing Signal Processing systems under hardware constraints, such as limited training epochs or evaluation times, presents a significant challenge. This task is often approached as a black-box optimization problem, as the performance metric, $ \mathcal{M}(\theta)$, is typically a complex function of the model's adjustable parameters $\Theta = \{\theta_1, \theta_2, ...,\theta_n\}$. The objective is to find the parameter configuration that maximizes a chosen metric (e.g., F1-score, classification accuracy). The problem  can be formulated as the following:

\begin{equation}
    \Theta^* = \arg\max_{\theta} \mathcal{M}(\Theta),
\end{equation}
where $ \theta^* $ is the optimal set of parameters that maximizes the performance metric $ \mathcal{M}(\theta) $.

\begin{algorithm}[t!]
    \caption{SignalLLM-Assisted Parameter Optimization} \label{SignalLLM4HP}
    \scriptsize
\begin{algorithmic}[1]
    \Require Initial parameter combinations $\theta_{\text{initial}}$, SP model $\mathcal{M}$, SignalLLM $\mathcal{L}$, Prompt $\mathcal{T}$, DE Algorithm $\mathcal{D}$
    \Ensure Optimized parameters $\theta^*$

    \State Initialize the solution-score pool $\mathcal{P}$ with pairs $(\theta, \mathcal{M}(\theta))$ for all $\theta \in \theta_{\text{initial}}$

    \For{$i \gets 1$ to $N$}
        \If{$i \pmod 2 \neq 0$} 

            \State Construct $ \mathcal{T} $ to guide SignalLLM reasoning
            \State Provide $ \mathcal{P} $ and $ \mathcal{T} $ as input to SignalLLM $ \mathcal{L} $
            \State SignalLLM analyzes $ \mathcal{P} $ and generates new combinations $ \theta_{\text{new}} $
            
        \Else 
            \State Generate new parameter combinations with DE Algorithm: $\theta_{\text{new}} \gets \mathcal{D}(\mathcal{P})$
        \EndIf
        
        \State Evaluate the performance score $s_{\text{new}} = \mathcal{M}(\theta_{\text{new}})$
        \State Update the pool $\mathcal{P} \gets \mathcal{P} \cup \{(\theta_{\text{new}}, s_{\text{new}})\}$
    \EndFor

    \State \Return $\theta^* \gets \arg\max_{(\theta, s) \in \mathcal{P}} s$
\end{algorithmic}
\end{algorithm}
Traditional optimization methods often require extensive evaluations, which can be computationally expensive and time-consuming. However, in resource-constrained environments, the number of evaluations must be minimized to ensure timely and efficient optimization. 

Our framework aims to addresse the complete spectrum of tunable parameters in SP tasks by this module, where SignalLLM offers a solution to this challenge by leveraging pre-trained knowledge and reasoning capabilities of LLM to optimize models with fewer evaluations. In this module, as formalized in Algorithm \ref{SignalLLM4HP}, LLM acts as an optimizer for SP models by generating and evaluating different configurations of hyperparameters. This capability is particularly useful in scenarios where traditional optimization methods struggle due to the complexity of the task or the lack of labeled data. By leveraging its ability to reason about statistical patterns, the LLM identifies optimal hyperparameter settings that improve model performance. The optimization process is guided by the LLM's understanding of the task's requirements and constraints, enabling it to explore the hyperparameter space more efficiently than conventional methods. This results in faster convergence and better performance, especially in tasks with high-dimensional or non-linear characteristics \cite{fang2025spatiotemporal}. 

To solve the optimization problem, we employ SignalLLM as an intelligent assistant that leverages its reasoning capabilities to suggest promising hyperparameter combinations. The optimization process involves the following steps:

\noindent\textit{i. Initialization:} Start with an initial set of hyperparameter combinations $ \theta_{\text{initial}} $, which are either randomly generated or based on prior knowledge. Evaluate the performance of each combination using the SP model and record the results in a Solution-Score Pairs Pool $ \mathcal{P} $.

\noindent\textit{ii. Solution-Score Pairs Pool:} The pool $ \mathcal{P} $ is a collection of previously evaluated hyperparameter combinations and their corresponding performance metrics:
     \begin{equation}
         \mathcal{P} = \{ (\theta_i, \mathcal{M}(\theta_i)) \mid i = 1, \dots, N \}
     \end{equation}
    This pool serves as a part of input to SignalLLM, enabling the model to analyze historical performance data and identify trends.

\noindent\textit{iii. SignalLLM Reasoning: }SignalLLM is provided with the Solution-Score Pairs Pool $ \mathcal{P} $ and a prompt $ \mathcal{T} $ that instructs the model to analyze the data and generate new hyperparameter combinations. The prompt $ \mathcal{T} $ includes: Description of the task,  Summary of the current best hyperparameters, Instructions to identify trends and gaps. SignalLLM uses its reasoning capabilities to analyze the pool and generate new combinations $ \theta_{\text{new}} $. Furthermore, to mitigate the numerical instability arising from the diversity of LLM responses, we alternate between the agent-based method and traditional differential evolution ~\cite{qin2008differential} in practical deployment.

\noindent\textit{iv. Evaluation and Feedback:} The new hyperparameter combinations $ \theta_{\text{new}} $ are evaluated using the SP model, and their performance metrics are recorded. The Solution-Score Pairs Pool $ \mathcal{P} $ is updated with the new results. This process is repeated iteratively, with SignalLLM continuously refining its suggestions based on the updated pool.

The use of SignalLLM for hyperparameter tuning offers several advantages, including data-driven reasoning, where SignalLLM analyzes historical performance data to make informed decisions about new hyperparameter combinations, reducing reliance on random search or trial-and-error methods. Additionally, SignalLLM excels at trend discovery, identifying subtle patterns and correlations between specific hyperparameters and high performance metrics, which leads to more effective optimization. Furthermore, by leveraging the Solution-Score Pairs Pool, SignalLLM achieves a balanced exploration-exploitation trade-off, exploring new combinations while focusing on those that have shown promise in the past, resulting in a more efficient and targeted optimization process.


\subsubsection{Parameter Transfer from Pre-Trained LLM (LLMs' Parameter-Level Support for SP Model)}
Pre-trained models offer a promising solution to the challenges of data scarcity and limited generalization in traditional DL-based SP methods. By leveraging prior knowledge, they enable efficient adaptation to complex SP tasks with minimal data and training, making them particularly valuable in resource-constrained settings. A capable SP agent should be able to invoke such models to complete tasks efficiently and robustly.
Inspired by recent work demonstrating that frozen pre-trained transformers from natural language processing can act as 
computation engines for time series analysis~\cite{NEURIPS2023_86c17de0}, SignalLLM adopts a parameter-level support strategy for specific SP modeling tasks. This approach refrains from altering the core self-attention and feedforward layers of the pre-trained transformer blocks, thereby preserving the rich, generalized features learned from vast datasets. To adapt the model for a given SP task, only a minimal set of parameters are fine-tuned. 
This includes redesigning the input embedding layer to project SP data into the required dimensions of the pre-trained model and fine-tuning the positional embeddings and layer normalization layers. 
Freezing the core transformer blocks while selectively training a small number of parameters leverages the weights from the model trained on broad datasets as a powerful starting point, enabling the fine-tuning process to find an effective solution efficiently, even in low-data regimes.


\section{Case Study and Experiments}


The evaluation of SignalLLM's capabilities requires systematic examination across fundamental signal processing domains and operational constraints. As detailed in Table~\ref{task_table}, to validate the effectiveness of SignalLLM, we conducted a series of experiments across multiple data modalities (e.g., text, sequential data, and images) under various learning scenario (e.g. few-shot, and fine-tuning settings) and various system constraints (from data scarcity to hardware limitations).  These experiments are designed to cover a wide range of SP tasks, encompassing key areas such as signal transmission, recognition, and perception. Specifically, the tasks included \textit{Few-Shot Radar Target Detection}, \textit{Zero-Shot Human Activity Recognition}, \textit{Text Signal Source Coding}, \textit{Handcrafted Feature Optimization}, and \textit{Modulated Signal Recognition}. 

\begin{table*}[h]
    \setlength{\tabcolsep}{2pt}

\centering
\caption{Rationale for case study selection. The chosen tasks are designed to evaluate SignalLLM across diverse core challenges, representative domains, and evaluation focuses to ensure a comprehensive assessment.}
\label{task_table}
\begin{tabular}{p{5cm}p{4cm}p{4cm}p{4cm}}
\hline
\textbf{Task} & \textbf{Core Challenge} & \textbf{Representative Domain} & \textbf{Evaluation Focus} \\
\hline
Few-Shot Radar Target Detection & Minimal training samples  & Radar Signal Processing & Few-shot adaptation \\
Zero-Shot Human Activity Recognition & Few prior target knowledge & Multi-Channel Signal Processing & Cross-domain generalization \\
Text Signal Source Coding & Lossless compression efficiency & Semantic communication & Context-aware coding \\
Handcrafted Feature Optimization & Poor flexibility & Non-convex optimization & Automated tuning \\
Modulated Signal Recognition & Resource-Limited Conditions & Wireless Signal Processing & Recognition accuracy\\
\hline
\end{tabular}
\end{table*}

The case study approach was chosen to provide a comprehensive evaluation of SignalLLM's capabilities in real-world scenarios. The selected tasks exemplify practical SP challenges where traditional methods require either extensive domain expertise (radar detection), specialized codes (text compression), or iterative manual tuning (hyperparameter optimization), where, we can demonstrate the model's adaptability, generalization, and performance under different conditions. This approach allows us to highlight the strengths of SignalLLM in handling diverse and complex SP challenges, while also identifying potential areas for improvement. By demonstrating SignalLLM's effectiveness across these diverse scenarios, we establish its potential as a general-purpose framework for next-generation intelligent signal processing systems.

\subsection{Datasets and Evaluation Metrics}
The evaluation framework employs standardized datasets and domain-specific metrics across five core signal processing tasks.

\subsubsection{Radar Target Detection} 

Radar target detection experiments are conducted using the IPIX database~\cite{Haykin_IPIX_Radar_Database}, a well-established resource in the field of maritime radar detection. 
Each scenario spans 131 seconds and encompasses 14 range cells, with 131,072 samples per cell (1 kHz sampling). 
The primary cells in these datasets contain the target returns, while the designated clutter cells provide observations of sea clutter. The secondary cells, which are adjacent to the primary cells, exhibit unique characteristics that distinguish them from other cells. The IPIX database is particularly valuable due to its detailed documentation of environmental conditions, such as wind speed and significant wave height, as well as the precise localization of targets within the primary cells. 3 datasets under different environmental conditions are selected from the database to evaluate SignalLLM. For the evaluation metrics, the F1-score and accuracy (ACC) are employed. The F1-score, defined as the harmonic mean of precision and recall, provides a balanced measure of a model’s performance in identifying both positive (target) and negative (clutter) instances, while accuracy quantifies the proportion of correctly classified samples relative to the total number of samples.

\subsubsection{Human Activity Recognition} 

We evaluate SignalLLM’s ability to classify human activities without task-specific training data, leveraging the Smartphone-Based Recognition of Human Activities and Postural Transitions Dataset~\cite{misc_smartphone-based_recognition_of_human_activities_and_postural_transitions_341}. The dataset contains raw inertial measurement unit (IMU) signals, including 3-axial linear acceleration and 3-axial angular velocity, sampled at 50Hz via smartphone accelerometers and gyroscopes. Twelve activities are originally annotated, spanning dynamic motions (e.g., walking) and static postures (e.g., lying). For the evaluation metrics, ACC is chosen.
\subsubsection{Text Signal Source Coding} 
Text signal coding experiments employ the proceedings of the European Parliament corpus~\cite{papineni-etal-2002-bleu}, which consists of around 2.0 million sentences and 53 million words. We use the first 90,000 sentences (2,175,972 tokens) grouped into 15-sentence blocks.  In this case, compression efficiency (CE) serves as the primary metric, defined as the ratio between original and compressed data sizes. 
 
\subsubsection{Handcrafted Feature Optimization} 
In this experiment, we investigate the ability of SignalLLM to assist in the optimization of handcrafted features. To comprehensively validate its effectiveness, we utilize the full suite of datasets from the IPIX radar database~\cite{Haykin_IPIX_Radar_Database}.
The evaluation is conducted in terms of both model score performance $S$ and F1-score performance, where the previous metrics is defined as:
\begin{equation}
S = P_d + \alpha \cdot \left(1 - P_{fa}\right),
\end{equation}
where $ P_d $ denotes the probability of detection, and $ P_{fa} $ represents the probability of false alarm. The hyperparameter $ \alpha $ (constrained by $ \alpha \gg 1 $, with $ \alpha = 10 $ in this study) acts as a penalty term to prioritize the suppression of false alarms. 
\subsubsection{Modulated Signal Recognition Under Resource-Limited Conditions}
In this study, we evaluate SignalLLM on the widely adopted RadioML 2016.10a dataset~\cite{zhang2021efficient}, which contains 11 classes of modulated signals collected under various channel conditions. To simulate moderately noisy scenarios, we select signal samples corresponding to signal-to-noise ratios (SNRs) of 0dB, 8dB, and 16 dB. The dataset is split into training, validation, and test sets with a ratio of 6:2:2, ensuring a balanced evaluation across different data splits. All models are trained for 20 epochs with a batch size of 256. Classification accuracy is adopted as the primary evaluation metric throughout our experiments.

\subsection{SP task Requirements, SignalLLM Reaction, and Baseline Methods}  

\subsubsection{Few-Shot Radar Target Detection} 
\begin{figure}[t] 
\centering
    \includegraphics[width=0.5\textwidth]{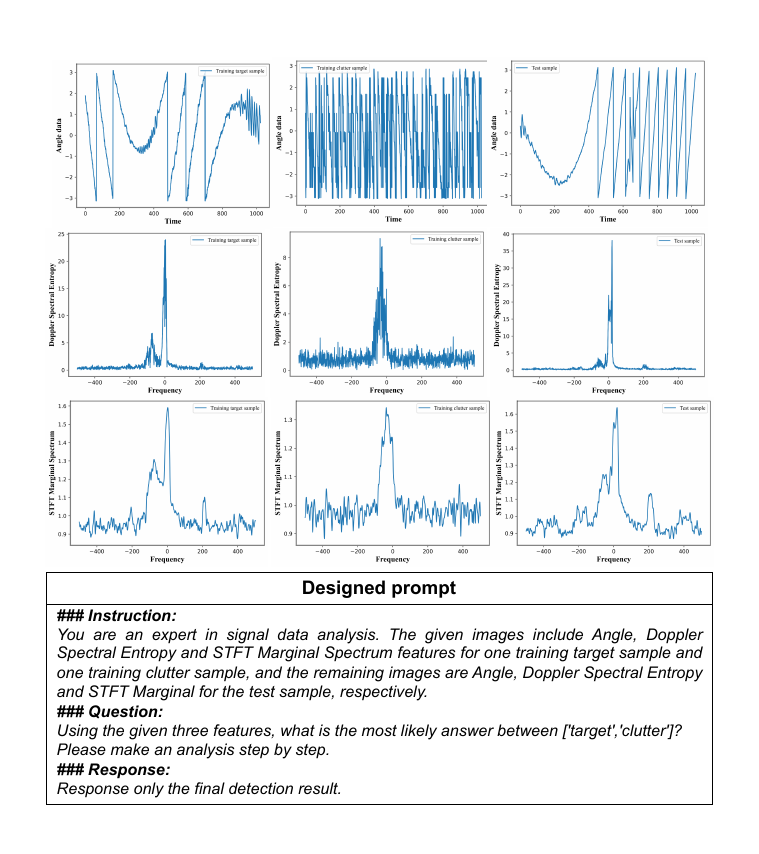}
        \caption{Illustration of a multimodal prompt used in few-shot radar target detection. The visual prompt (top) provides representative radar data samples to guide model perception, while the text prompt (bottom) offers task-specific instructions or semantic priors to complement the visual modality and enable cross-modal reasoning. }
        \label{f_propmt}
\end{figure}

In this task, SignalLLM is tasked with performing radar target detection using only two training samples, one representing the target and the other representing clutter. To enhance its discriminative capability, SignalLLM first analyzes the radar signals by extracting features such as Angle, Doppler Spectral Entropy, and the STFT Marginal Spectrum. It then constructs a structured prompt comprising four components: Visualization of Features, Instruction, Question, and Response, and leverages GPT-4o~\cite{openai2024hello} for cross-modal reasoning. The automatically generated prompt is illustrated in Fig.~\ref{f_propmt}. Regarding the compared methods, we select two state-of-the-art handcrafted detection algorithms~\cite{zhou2019decision,li2019svm}, both trained with sufficient data (30\% of the dataset used for training).

\subsubsection{Zero-Shot Human Activity Recognition} 
 For human activity recognition, we follow the task requirements proposed in previous work~\cite{an2024iot}, focusing on two standard classification tasks: a binary classification between WALKING and STANDING, and a ternary classification among the LYING, WALKING UPSTAIRS, and LIE TO SIT transitions. First, SignalLLM retrieves relevant descriptions of the three activities from a knowledge base. It then constructs a structured prompt consisting of four components: signal visualizations, instructions (based on the retrieved knowledge), questions, and responses. This prompt is subsequently used to facilitate cross-modal reasoning via GPT-4o~\cite{openai2024hello}. To evaluate the effectiveness of SignalLLM, we compare it with an agent-based baseline, IoT-LLM~\cite{an2024iot}, which employs carefully designed prompts to perform HAR in a zero-shot setting.
 
\subsubsection{Text Signal Source Coding} In this task, we employ SignalLLM to perform text-based signal source coding under sufficient computational resources, aiming to achieve maximal compression. Specifically, SignalLLM first adopts the compression strategy outlined in Algorithm~\ref{LLM4code}, and then invokes the pre-trained GPT-2 model to execute the compression task. For comparison, we adopt three traditional handcrafted coding algorithms~\cite{xie2021deep} that do not rely on pre-trained models. In contrast, SignalLLM leverages LLMs to generate compact, semantically guided representations for more adaptive compression.

\subsubsection{Handcrafted Feature Optimization} 

In this task, we employ three handcrafted features: Frequency Peak-to-Average Ratio (FPAR) with an adjustable frequency range hyperparameter $\theta_1$, Short-Time Fourier Transform Mean (STFTM) with a tunable neighborhood ratio hyperparameter $\theta_2$, and Time Information Entropy (TIE) with a configurable interval number hyperparameter $\theta_3$. These features are used in conjunction with a linear regression classifier to distinguish signals from different classes. SignalLLM first extracts the three features from the radar signals and then optimizes their corresponding hyperparameters. The hyperparameters $\theta_1$, $\theta_2$, and $\theta_3$ are tuned via iterative evaluation of solution-score pairs, defined as $\mathcal{P} = { (\theta_i, \mathcal{M}(\theta_i)) }$, with each trial strictly constrained to 100 optimization iterations.
We compare SignalLLM against two representative handcrafted optimization algorithms: Differential Evolution (DE)\cite{qin2008differential} and Simulated Annealing (SA)\cite{van1987simulated}. In contrast, SignalLLM leverages the powerful reasoning capabilities of LLMs to perform flexible and context-aware hyperparameter optimization.
\subsubsection{Modulated Signal Recognition Under Resource-Limited Conditions}
In this task, SignalLLM is applied to modulated signal recognition under resource-constrained conditions. It first transforms the input signal using the Short-Time Fourier Transform (STFT), then fine-tunes a pre-trained Convolutional Neural Network (CNN) to adapt to the limited number of training steps. For comparison, two manually designed methods are included~\cite{zhang2021efficient,xu2020spatiotemporal}.

\subsection{Experimental Results}


\subsubsection{SignalLLM-Asssisted Few-Shot Radar Target Detection}
\begin{table}[t]
\centering
\caption{Detection results on the given three datasets. We express all evaluative metrics as percentages. The best and runner-up results are highlighted with \textbf{bold} and \underline{underline}, respectively.} \label{detect}
\setlength{\tabcolsep}{1.0mm}{
\begin{tabular}{ccccc}
\hline
& Metrics & Li \textit{et al.}~\cite{li2019svm} & Zhou \textit{et al.}~\cite{zhou2019decision}& SignalLLM \\ \hline
\multirow{2}{*}{\#17} & ACC & \underline{87.75} & 87.57 & \textbf{88.36} \\ \cline{2-5} 
& F1-score & \underline{88.10} & 86.78 & \textbf{88.36} \\ \hline
\multirow{2}{*}{\#310} & ACC & 82.95 & \underline{91.39} & \textbf{93.36} \\ \cline{2-5} 
& F1-score & 80.46 & \underline{90.88} & \textbf{92.52} \\ \hline
\multirow{2}{*}{\#311} & ACC & 95.61 & \underline{96.45} & \textbf{97.36} \\ \cline{2-5} 
& F1-score & 95.51 & \underline{96.38} & \textbf{97.36} \\ \hline
\end{tabular}
}
\end{table}
As shown in Table~\ref{detect}, SignalLLM consistently outperforms SOTA manually designed detection methods across all three datasets in terms of both accuracy and F1-score. This consistent superiority highlights the robustness and reliability of SignalLLM in radar target detection. Moreover, the experimental results demonstrate its strong efficacy under limited training data conditions, making it particularly suitable for data-scarce scenarios.

\subsubsection{SignalLLM-Assisted Zero-Shot Human Activity Recognition}

\begin{table}[t]
\caption{Human Activity Recognition within our framework on two tasks. We express all evaluative metrics as percentages. The best and runner-up results are highlighted with \textbf{bold} and \underline{underline}, respectively.}
\begin{center}
\scalebox{1}{
    \label{ablation}
    \begin{tabular}{c|cc}
    \toprule
    \multirow{2}{*}{ Method} & \multicolumn{2}{c}{\textbf{IoT tasks}~(ACC$\uparrow$)} \\
    \cmidrule{2-3}
    \multirow{2}{*}{} & HAR-2cls & HAR-3cls \\
    \midrule
    GPT-4 Baseline & \underline{77.3} & 43.3 \\
    \midrule
    IoT-LLM~\cite{an2024iot} & \textbf{100.0} & \underline{87.8} \\
    \midrule
    SignalLLM & \textbf{100.0} & \textbf{92.5} \\
    \bottomrule
    \end{tabular}
}
\end{center}
\label{HR}
\end{table}

As shown in Table~\ref{HR}, SignalLLM consistently outperforms existing agent-based methods across both datasets, demonstrating its remarkable zero-shot learning capability in SP tasks.





\subsubsection{SignalLLM-Assisted Text Signal Source Coding}
\begin{table}[t]
\caption{Compression efficiency of different coding methods. The best and runner-up results are highlighted with \textbf{bold} and \underline{underline}, respectively.}
\centering
\setlength{\tabcolsep}{1.4mm}{
\begin{tabular}{ccccccc}
\hline
   & Huffman & 5-bit & Brotli & $K=10$ & $K=20$ & $K=40$ \\ \hline
CE &   1.87      &  1.39     &   3.07     &  8.08   &  \underline{8.50} &  \textbf{8.97} \\ \hline
\end{tabular}
 \label{code_CE}
}
\end{table}

As shown in Table~\ref{code_CE}, the SignalLLM-assisted method achieved significantly higher CE, particularly as the selected prior context length ($K$) increased. With $K = 40$, the CE reached 8.97, significantly surpassing traditional methods.

\subsubsection{SignalLLM-Assisted Handcrafted Feature Optimization}
\begin{figure}[!t]
    \centering{        \includegraphics[width=0.47\linewidth]{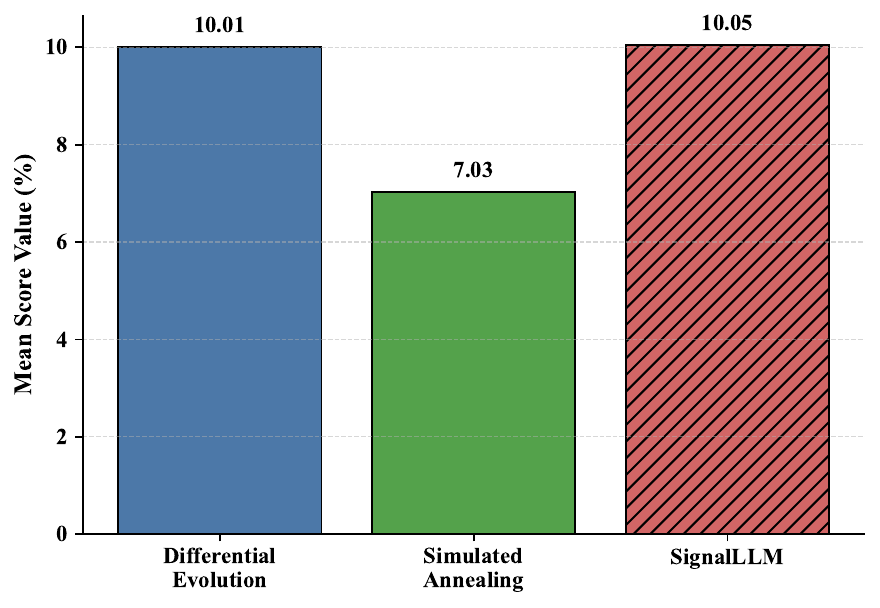}
        }
    \hfill{
        \includegraphics[width=0.47\linewidth]{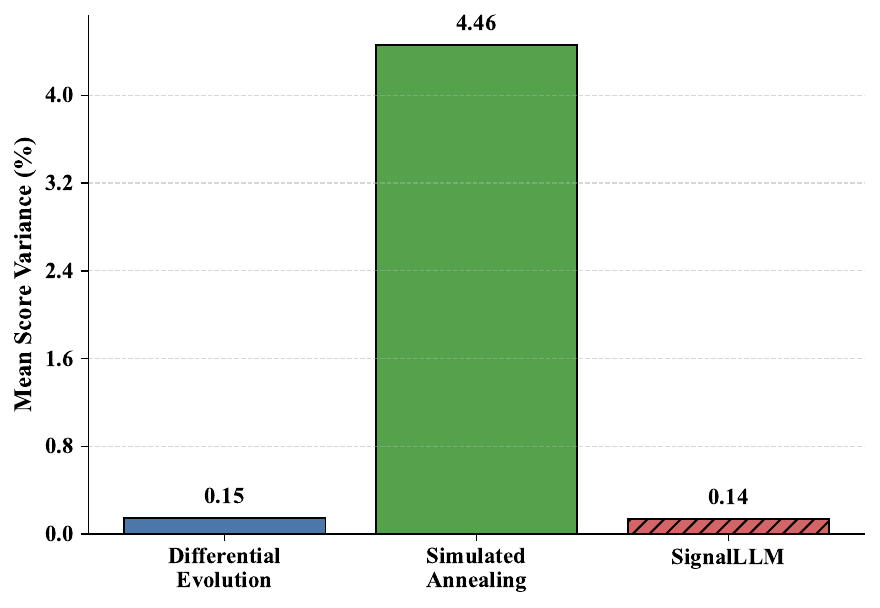}
        }
    \vfill
    {
        \includegraphics[width=0.47\linewidth]{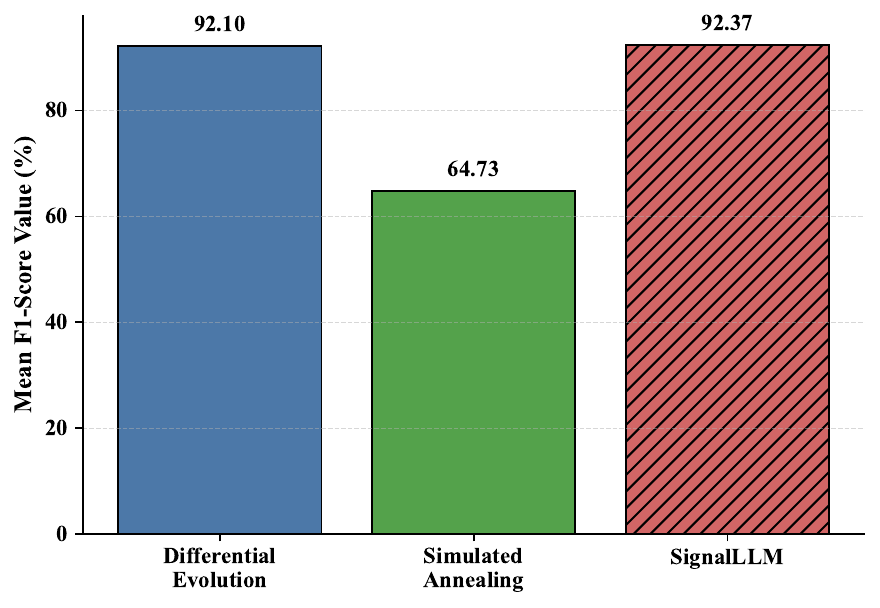}
        }
    \hfill
    {
        \includegraphics[width=0.47\linewidth]{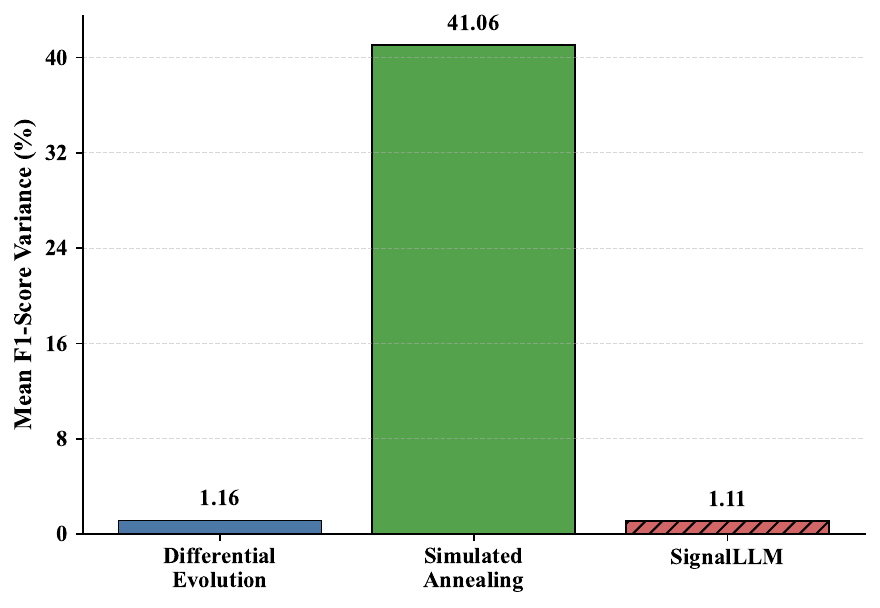}
        }
    \caption{Model score and F1-score Performance comparisons on benchmark datasets.}
    \label{fig:Optimize}
\end{figure}

As illustrated in Fig.\ref{fig:Optimize}, the experimental results validate the effectiveness of SignalLLM in optimizing handcrafted features for signal classification tasks. Under a constrained budget of 100 optimization trials, each method was evaluated over 20 independent runs to ensure statistical reliability. SignalLLM consistently outperforms manually designed optimization algorithms, including Differential Evolution (DE)~\cite{qin2008differential} and Simulated Annealing (SA)~\cite{van1987simulated}, by achieving both higher average performance and lower performance variance. Notably, it achieves state-of-the-art (SOTA) results across nearly all evaluated datasets.

\subsubsection{SignalLLM-Assisted Modulated Signal Recognition Under Resource-Limited Conditions}
\begin{table}[t]
\caption{Mean classification accuracies under $0$, $8$, and $16$ dB SNR. Metrics are given in percentages, with the best and second-best results shown in bold and \underline{underline}, respectively. ACC($n$) denotes accuracy at $n$ dB SNR.} 
\centering
\setlength{\tabcolsep}{1.4mm}{
\begin{tabular}{ccccc}
\hline
   & PET-CGDNN~\cite{zhang2021efficient} & FEA\_T~\cite{xu2020spatiotemporal} & SignalLLM \\ \hline
 ACC(0) &    53.86     &  \underline{65.40}    &  \textbf{80.41}    \\ \hline 
 ACC(8) &    53.16   &   \underline{71.66}    &  \textbf{81.59} \\ \hline
 ACC(16) &    59.56     &  \underline{70.97}    &  \textbf{84.01}  \\ \hline
\end{tabular}
\label{ACC}
}
\end{table}

 The classification accuracy of SignalLLM and other baseline methods under 0, 8, and 16dB SNR are shown in Table~\ref{ACC}. Notably, the SignalLLM approach exhibits robust classification accuracy under constrained training resources, outperforming the manually designed methods by a considerable margin.

\section{Discussion}
In general, SignalLLM marks a major and impactful advancement in signal processing (SP) research, demonstrating robust and substantial improvements over both manually designed and agent-based methods. Its strong performance across five complex SP tasks under extreme constraints highlights the significant potential of LLMs to enhance traditional SP algorithms and automate complex SP workflows that have historically required extensive human effort. 

Importantly, our work provides the first empirical evidence that planning across diverse signal processing (SP) action spaces via agent-based workflows can produce solutions that surpass human-designed heuristics. This finding opens a promising research direction: instead of relying on static or handcrafted procedures, LLM-based agents can dynamically orchestrate heterogeneous tools, pre-trained models, and actions, thereby uncovering more effective and generalizable SP pipelines than those crafted by domain experts.

Nonetheless, several limitations remain. Since the planning stage of SignalLLM requires appropriate decision-making across a wide range of action spaces, tools, and pre-trained models, the tasks evaluated thus far represent only a narrow subset of the broader SP domain. Although SignalLLM exhibits impressive capabilities at its current stage, effectively selecting suitable action spaces, tools, and pre-trained models, its performance on more complex SP problems remains to be further validated. In particular, SignalLLM does not yet consistently achieve strong results across diverse and challenging scenarios. To further improve its effectiveness, we suggest incorporating more advanced RAG techniques and agent memory mechanisms~\cite{xu2025mem}, or fine-tuning the model with reinforcement learning methods~\cite{sun2024llm}, thereby enabling more informed and adaptive decision-making.

\section{Conclusion}
This paper presents SignalLLM, a general-purpose LLM-based agent framework for automating and generalizing diverse SP tasks. By combining structured task decomposition, adaptive planning with RAG, solution refining module, and hybrid task execution module, SignalLLM mitigates key challenges in both traditional and prior LLM-based SP approaches. Empirical results across representative SP benchmarks demonstrate its superior performance over traditional SP methods and existing agent-based methods. This approach not only reduces the reliance on manual intervention, but also opens new avenues for solving complex SP challenges in a fully automated and intelligent manner.

In the future, we plan to expand the benchmark suite to cover a broader spectrum of SP scenarios, including audio, biomedical, and geophysical signals. Further exploration of lightweight model variants, lower API costs, and real-time applications will also enhance its practicality in resource-constrained environments.








\bibliographystyle{IEEEtran}
\bibliography{bibtex/bib/IEEEexample}
\clearpage
\appendices
\end{document}